\documentclass[11pt]{article}

\usepackage{acl}
\usepackage{amssymb}
\usepackage{times}
\usepackage{algorithm}
\usepackage{algorithmic}
\usepackage{latexsym}
\usepackage{subcaption}
\usepackage{amsmath}
\usepackage{booktabs}
\usepackage{multirow}
\usepackage{booktabs}
\usepackage[table]{xcolor}
\usepackage{makecell}
\usepackage[T1]{fontenc}
\usepackage[utf8]{inputenc}

\usepackage{microtype}

\usepackage{inconsolata}

\usepackage{graphicx}

\title{GeoRA: Geometry-Aware Low-Rank Adaptation for RLVR}

\author{
  \textbf{Jiaying Zhang}\textsuperscript{1,2\dag*}, % 第一行加粗
  \textbf{Lei Shi}\textsuperscript{1*\ddag},          % 第二行加粗
  \textbf{Jiguo Li}\textsuperscript{1},  \\      % 后面依次加粗...
  \textbf{Jun Xu}\textsuperscript{1\ddag},
  \textbf{Jiuchong Gao}\textsuperscript{1\ddag},
  \textbf{Jinghua Hao}\textsuperscript{1},
  \textbf{Renqing He}\textsuperscript{1}
  \\
  \textsuperscript{1}Meituan, \textsuperscript{2}Peking University
    \\
    {\texttt{zhangjy2002@stu.pku.edu.cn}}
\\
    {\texttt{\{shilei74, xujun58, gaojiuchong\}@meituan.com}}
}

\begin{document}
\maketitle
\insert\footins{\noindent\footnotesize\textsuperscript{\dag} Work done during internship at Meituan.} 
\insert\footins{\noindent\footnotesize\textsuperscript{\ddag} Corresponding authors.} 
\insert\footins{\noindent\footnotesize\textsuperscript{*} Equal contribution.} 

\begin{abstract}
Reinforcement Learning with Verifiable Rewards (RLVR) is a key paradigm for improving large-scale reasoning models. Unlike supervised fine-tuning (SFT), RLVR exhibits distinct optimization dynamics and is sensitive to the preservation of pre-trained geometric structures. However, existing parameter-efficient methods face key limitations in this regime. Low-rank adaptation methods, such as PiSSA, are primarily designed for Supervised Fine-Tuning (SFT) and do not account for the distinct optimization dynamics and geometric structures of RLVR. Conversely, directly fine-tuning the unstructured sparse parameter subspace favored by RLVR encounters efficiency bottlenecks on modern hardware. To address these challenges, we propose GeoRA (Geometry-Aware Low-Rank Adaptation), a low-rank adaptation method tailored for RLVR. Specifically, GeoRA exploits the anisotropic and compressible structure of RL update subspace, and extracts its principal directions via Singular Value Decomposition (SVD) to initialize low-rank adapters, while freezing residual components as a structural anchor during training. This design preserves the pre-trained structure and enables efficient dense computation. Experiments on Qwen and Llama models from 1.5B to 32B parameters show that GeoRA consistently outperforms strong low-rank baselines across RLVR settings in mathematics, medicine, and coding, while showing stronger generalization and less forgetting on out-of-domain tasks.

%“基于可验证奖励的强化学习（RLVR）是提升大规模推理模型能力的关键范式。与监督微调（SFT）不同，RLVR 展现出截然不同的优化动力学，其对预训练几何结构的保持高度敏感。然而，现有的参数高效方法在这一范式下面临着关键的局限性。一方面，像 PiSSA 和 MiLoRA 这样的低秩适配方法主要面向监督微调设计，未能考虑到 RLVR 独特的优化动力学与几何结构；直接将它们应用于 RLVR 往往会导致谱塌缩和训练不稳定。另一方面，直接去微调那些受 RLVR 青睐的非结构化稀疏参数子空间，又会在现代硬件上遭遇效率瓶颈。为解决这些挑战，我们提出了 GeoRA（几何感知低秩适配），一种专为 RLVR 定制的低秩适配方法。具体而言，GeoRA 利用了 RL 更新子空间的各向异性与可压缩结构，并通过奇异值分解（SVD）提取其主方向来初始化低秩适配器。在 RLVR 训练期间，残差分量被冻结作为结构锚点。该设计既保留了预训练结构，又实现了高效的稠密计算。在参数规模从 1.5B 到 32B 的 Qwen 和 Llama 模型上的实验表明，GeoRA 在数学、医学和代码等 RLVR 场景中持续优于强有力的低秩基线方法，同时在域外任务上展现出更强的泛化能力和更少的遗忘。”
\end{abstract}

\section{Introduction}

Large reasoning models, represented by OpenAI-o1~\cite{openai2024openaio1card} and DeepSeek-R1~\cite{deepseekai2025deepseekr1incentivizingreasoningcapability}, have established Reinforcement Learning with Verifiable Rewards (RLVR) as a pivotal paradigm for unlocking complex reasoning capabilities. Unlike supervised fine-tuning (SFT), RLVR is better characterized as a constrained optimization process~\cite{5wu2025invisibleleashrlvrescape} that amplifies latent reasoning behaviors through reward-induced sampling bias~\cite{6yue2025doesreinforcementlearningreally,9zhao2025echochamberrlposttraining}. As a result, RLVR is particularly sensitive to update stability and its alignment with pre-trained representation geometry: overly aggressive updates can collapse behavior or degrade general capabilities. Empirically, substantial gains can emerge from modifying only a small fraction of parameters~\cite{mukherjee2025subnetworks}, and recent mechanistic studies further suggest that effective RLVR updates are geometrically biased toward preserving pre-trained structure~\cite{1offtheprincipals,2cai2025predictabilityreinforcementlearningdynamics}.

However, existing PEFT methods face key limitations in this regime. First, SFT-oriented low-rank adaptation methods suffer from a geometric mismatch with RLVR~\cite{yin2025evaluatingpeft_rlvr}. PiSSA~\cite{pissa}, for example, forces updates onto principal components, conflicting with RLVR's preferred subspace while protecting core features. Second, some sparse fine-tuning methods~\cite{mukherjee2025subnetworks,1offtheprincipals}, although more consistent with RLVR update patterns, struggle to achieve practical efficiency. Because modern hardware provides limited support for unstructured sparsity, these methods often fail to translate theoretical sparsity into real-world speedups, and may even introduce additional overhead.

To address these challenges, we introduce \textbf{GeoRA} (Geometry-Aware Low-Rank Adaptation), a low-rank adaptation framework tailored to RLVR. Our analysis shows that the effective RLVR update subspace is anisotropic and compressible rather than isotropic, exhibiting a structured low-rank form. Based on this observation, GeoRA extracts dominant trainable directions from a geometry-constrained subspace for initialization, while keeping residual components frozen as a structural anchor. In this way, GeoRA simultaneously aligns adaptation with RLVR-specific optimization geometry and preserves dense matrix computation, achieving both stable optimization and hardware-efficient training. To the best of our knowledge, GeoRA is the first geometry-aware low-rank adaptation framework explicitly designed for RLVR. Our contributions are summarized as follows:

\begin{itemize}
\item We propose GeoRA, a geometry-aware, low-rank, and parameter-efficient adaptation framework tailored to RLVR. By aligning low-rank adaptation with RLVR-specific optimization geometry while preserving dense computation, GeoRA overcomes the geometric mismatch of SFT-oriented low-rank methods and the efficiency bottleneck of sparse methods.
% 中文：我们提出GeoRA：一种面向 RLVR 的几何感知、低秩且参数高效的适配框架。GeoRA 在保留稠密计算的同时，将低秩适配与 RLVR 特有的优化几何对齐，从而缓解面向 SFT 的低秩方法的几何失配问题以及稀疏方法的效率瓶颈。

\item We show that the effective RL update subspace is directional and admits a compressible low-rank structure. GeoRA extracts dominant trainable directions via SVD within this subspace to initialize low-rank adapters, while a frozen residual component acts as a structural anchor to preserve pre-trained structure.
% 中文：我们表明有效的 RL 更新子空间具有方向性，并具有可压缩的低秩结构。GeoRA 在该子空间内通过 SVD 提取主导可训练方向来初始化低秩适配器，并以冻结的残差分量作为结构锚点以保留预训练结构。

\item Experiments on Qwen and Llama models from 1.5B to 32B parameters demonstrate that GeoRA improves training stability and consistently outperforms strong low-rank baselines across diverse RLVR settings, while showing stronger generalization and less forgetting on out-of-domain tasks.
% 中文：在 Qwen 与 Llama 系列 1.5B 到 32B 模型上的实验表明，GeoRA 在提升训练稳定性的同时，在多种 RLVR 场景下持续优于强有力的低秩基线方法，并在域外任务上表现出更强的泛化能力和更少的遗忘。
\end{itemize}

\begin{figure*}[t]
  \centering
  \includegraphics[width=0.9\textwidth]{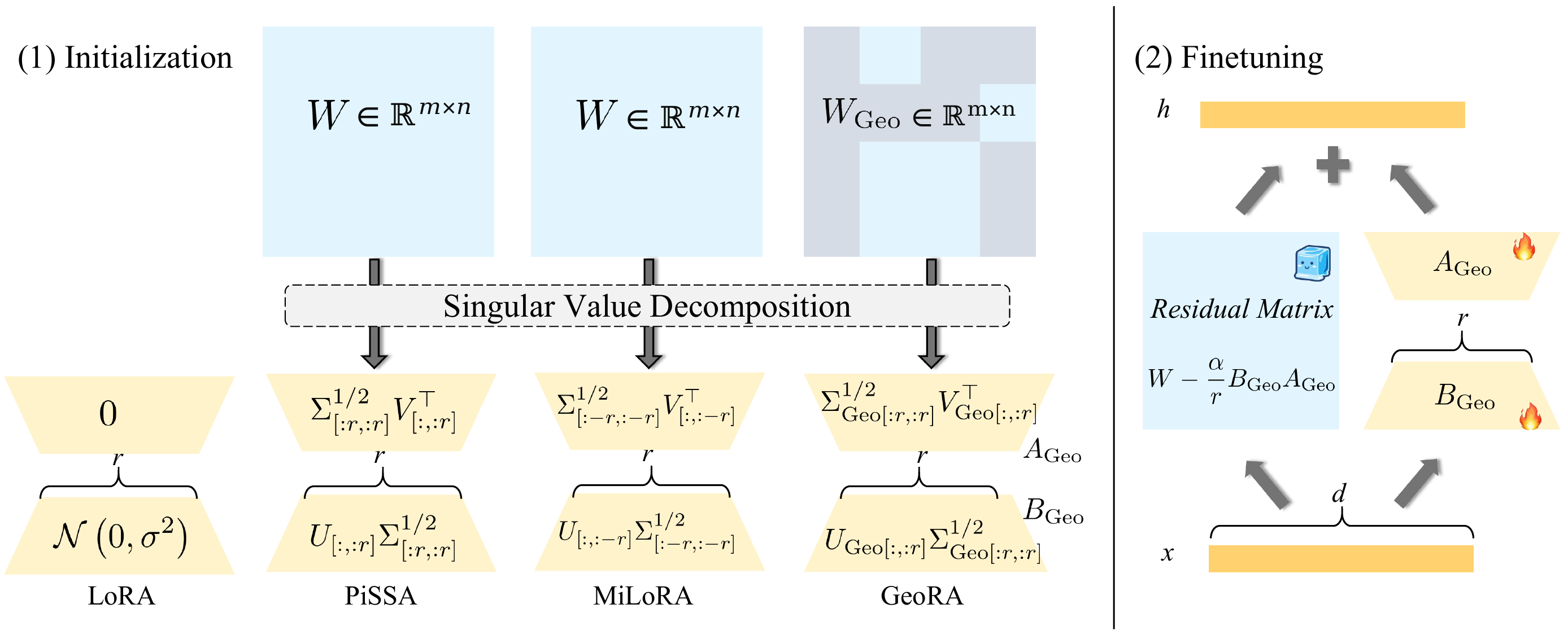} % 建议宽度设为 0.8 左右，避免铺满全屏
  \caption{Comparison of adapter initialization and forward architectures. \textbf{LoRA} applies low-rank adaptation on the original weight matrix $W$ with standard initialization, while \textbf{PiSSA} initializes adapters from the principal components of $W$. In contrast, \textbf{GeoRA} initializes from a geometry-constrained matrix $W_{\text{Geo}}$ (a different adaptation target than $W$). Its forward pass incorporates a frozen Residual Matrix in parallel with the trainable adapter to act as a stability anchor for principal components.}
  \label{fig:init}
\end{figure*}
% 中文：图~\ref{fig:init} 对比了不同适配器初始化与前向结构：LoRA 在原始权重矩阵 $W$ 上做低秩适配并采用标准初始化，PiSSA 从 $W$ 的主成分初始化；而 GeoRA 则在几何约束矩阵 $W_{\text{Geo}}$ 上初始化（适配的目标矩阵不同于 $W$），并通过冻结残差分量提供主成分的稳定锚点。

\begin{figure*}[t]
  \centering
  \includegraphics[width=0.9\textwidth]{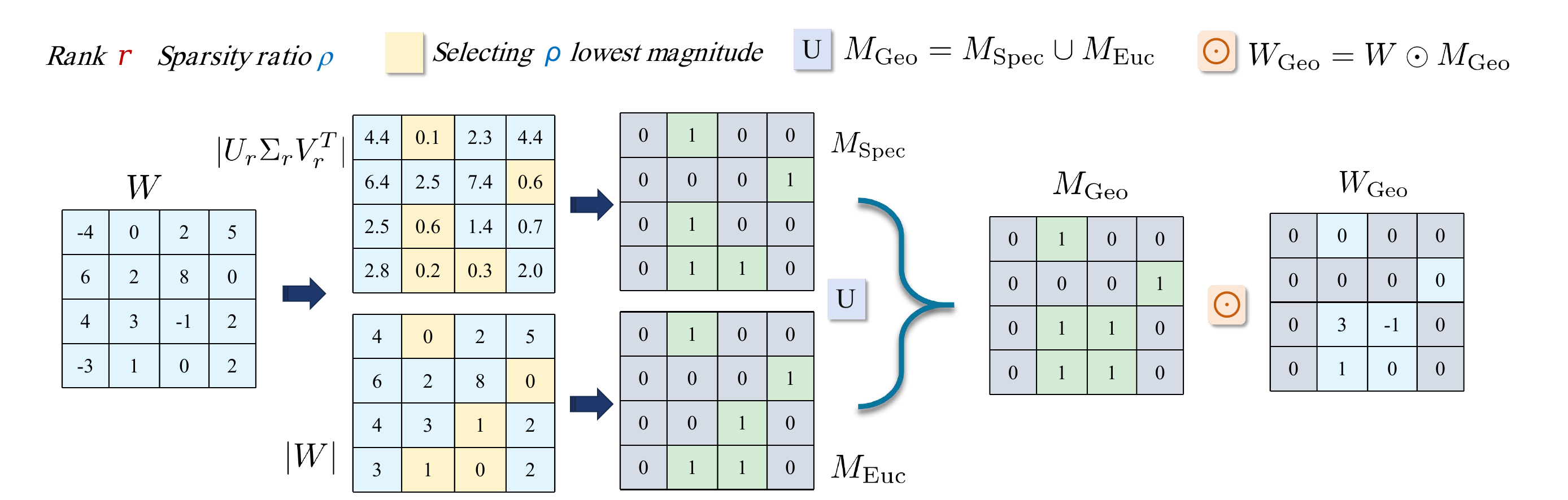} % 建议宽度设为 0.8 左右，避免铺满全屏
  \caption{Geometric Prior Construction via Masking. The process of generating $M_{\text{Geo}}$ by combining Spectral Priors (low-curvature regions) and Euclidean Priors (high-plasticity near-zero weights). The resulting $W_{\text{Geo}}$ isolates the most stable parameters for RL-native adaptation.}
  \label{fig:alignment_wide}
\end{figure*}
% 中文：图~\ref{fig:alignment_wide} 展示了掩码构造：谱先验用于避开高曲率主方向，欧氏先验用于选择高可塑性的近零权重区域，二者合并得到用于 RL 原生适配的稳定参数集合。

\section{Related Work}
\subsection{RLVR and Optimization Geometry}
RLVR replaces traditional reward models with deterministic verifiers (e.g., in math or coding)~\cite{rlsurvey,rl1,rl4}. Through outcome-based feedback, it incentivizes emergent reasoning behaviors like Chain-of-Thought (CoT), establishing itself as a core paradigm for enhancing LLM reasoning~\cite{rl6,rl8}.
% 中文：RLVR 用确定性的验证器（如数学/代码判题）替代传统奖励模型，通过基于结果的反馈激励链式思维等推理行为，已成为提升大模型推理能力的核心范式之一。

Recent mechanistic analyses have delineated a sharp dichotomy between supervised fine-tuning (SFT) and reinforcement learning with verifiable rewards (RLVR). While SFT primarily injects knowledge by modifying principal weight directions~\cite{12chu2025sftmemorizesrlgeneralizes,14jin2025rlpanaceamirageunderstanding}, RLVR is better characterized as a constrained optimization process~\cite{5wu2025invisibleleashrlvrescape} that amplifies latent reasoning behaviors via reward-induced sampling bias rather than introducing new capabilities~\cite{6yue2025doesreinforcementlearningreally,9zhao2025echochamberrlposttraining,15alam2025limitsgeneralizationrlvrcase}. Theoretically, these updates manifest ``off the principals,'' favoring low-magnitude directions orthogonal to pre-trained features~\cite{1offtheprincipals}, consistent with the rank-1 dominance observed in early training~\cite{2cai2025predictabilityreinforcementlearningdynamics}. However, this regime faces stability trade-offs. KL-regularization (``RL's Razor'') attempts to limit forgetting~\cite{4shenfeld2025rlsrazoronlinereinforcement} but can precipitate the ``Reasoning Boundary Paradox,'' where aggressive reward maximization collapses exploration diversity~\cite{7nguyen2025reasoningboundaryparadoxreinforcement}. Although capability degradation may be partially reversed via singular vector rotation~\cite{11jin2025rlfinetuninghealsood,14jin2025rlpanaceamirageunderstanding}, adaptation remains fundamentally constrained by the ``Invisible Leash,'' which enforces proximity to the pre-training manifold~\cite{5wu2025invisibleleashrlvrescape}.
Empirically, strong gains can emerge from updating only a small fraction of parameters, suggesting that RL fine-tuning often concentrates on small subnetworks~\cite{mukherjee2025subnetworks}.
% 中文：近期机制研究指出，SFT 与 RLVR 的训练动力学存在显著差异：SFT 倾向于通过修改主方向注入知识，而 RLVR 更像受约束的优化过程，主要通过奖励诱导的采样偏置放大已有推理能力；理论上更新偏向“远离主方向”的低幅值/低曲率子空间，并在早期呈现近似 rank-1 的主导结构。与此同时，KL 正则（RL's Razor）与探索多样性之间存在稳定性权衡，过强的奖励最大化可能触发推理边界坍塌；即便通过奇异向量旋转能部分修复退化，整体仍受“隐形牵绳”约束以贴近预训练流形。另外，实证结果显示只更新少量参数即可获得显著收益，提示 RL 微调常集中在小规模子网络上。

\subsection{PEFT and Spectral Priors}
To alleviate the computational demands of scaling LLMs, PEFT~\cite{peft1,peft2} has emerged as a key paradigm, minimizing memory overhead by updating only a fraction of parameters while matching full fine-tuning performance. Prevailing strategies include partial fine-tuning~\cite{peft11,peft12}, soft prompt tuning~\cite{peft21}, non-linear adapters~\cite{peft31}, low-rank adaptation~\cite{peft42}, and importance-aware methods like LIFT~\cite{lift}. As a practical paradigm, PEFT has also been effectively adapted to continual learning~\cite{liang2023prompts} and multi-task learning~\cite{liu2023hierarchical} scenarios.
% 中文：为降低大模型微调的计算与存储开销，参数高效微调（PEFT）通过仅更新少量参数来尽可能接近全参微调效果，典型方法包括部分微调、提示微调、非线性适配器、低秩适配以及如 LIFT 等重要性感知方法。作为一种实用范式，PEFT 也被有效拓展到持续学习与多任务学习等场景。

Among these, LoRA~\cite{lora} and its variants are the de facto standard, yet their initialization strategies often diverge based on spectral priors. For instance, PiSSA~\cite{pissa} allocates trainable parameters to principal singular components, imposing a strong inductive bias validated primarily in SFT. However, such SFT-oriented spectral priors can create a fundamental geometric mismatch in RLVR, whose optimization dynamics and effective update patterns differ markedly from SFT~\cite{yin2025evaluatingpeft_rlvr,1offtheprincipals,mukherjee2025subnetworks}. MiLoRA~\cite{milora} instead targets minor components, but does not explicitly account for RLVR-specific geometry and dynamics. Other variants like DoRA~\cite{dora}, AdaLoRA~\cite{adalora}, and VeRA~\cite{vera} focus on weight decomposition or budget allocation without integrating these RLVR-native geometric considerations. Building on these observations, GeoRA translates RLVR-specific mechanistic insights into an actionable PEFT paradigm.
% 中文：在 PEFT 中，LoRA 及其变体最为常用，但其初始化往往隐含不同的谱先验：例如 PiSSA 将可训练参数分配到主奇异分量，这一强归纳偏置主要在 SFT 语境下得到验证；但将此类面向 SFT 的谱先验直接用于 RLVR 时，往往会出现几何/动力学层面的结构性错配，因为 RLVR 的优化动力学与有效更新模式与 SFT 显著不同。MiLoRA 则选择尾部（minor）奇异分量，但并未显式对齐 RLVR 特有的几何/动力学约束。DoRA/AdaLoRA/VeRA 等更多聚焦权重分解或预算分配，同样未将 RLVR 相关的几何考虑系统性纳入。基于这些观察，我们提出 GeoRA，将 RLVR 相关的机制洞见转化为可操作的 PEFT 方案。

\section{Methodology}

\label{sec:method}

We introduce \textbf{GeoRA} (Geometry-Aware Low-Rank Adaptation), a low-rank adaptation framework tailored for RLVR. GeoRA emphasizes an RL-native, structured low-rank parameterization together with an explicit residual leash for stability. Comparing with classical PEFT baselines like LoRA and PiSSA, we first present the low-rank forward architecture and initialization (Figure~\ref{fig:init}), and then describe a lightweight masking instantiation that constructs the geometry-constrained matrix used for initialization (Figure~\ref{fig:alignment_wide}). 
% The complete pseudocode of our algorithm is provided in Appendix~\ref{app:algorithm}.
% 中文：我们提出 \textbf{GeoRA}：一个面向 RLVR 的 PEFT 框架。GeoRA 强调 RL 原生的结构化低秩参数化，并通过显式的残差牵绳保证稳定性。对比 LoRA 和 PiSSA 等面向 SFT 的 PEFT 基线，我们先介绍低秩前向结构与初始化（图~\ref{fig:init}），再说明用于实例化几何约束区域的掩码构造方式（图~\ref{fig:alignment_wide}）。

\subsection{Geometry-Aware Low-Rank Structure}

Let $W_{\text{Geo}}$ denote a geometry-constrained view of the pre-trained weight matrix $W$. Unlike standard LoRA which initializes adapters randomly (or with zero), GeoRA leverages the compressible structure within $W_{\text{Geo}}$ to derive a structured initialization. 
% 中文：记 $W_{\text{Geo}}$ 为预训练权重 $W$ 的几何约束视图。与标准 LoRA 采用随机（或零）初始化不同，GeoRA 利用 $W_{\text{Geo}}$ 内的可压缩结构导出结构化初始化。

We first perform Singular Value Decomposition (SVD) on the geometry-constrained matrix:
% 中文：我们首先对该几何约束矩阵执行奇异值分解（SVD）：
\begin{equation}
    W_{\text{Geo}} = U_{\text{Geo}} \Sigma_{\text{Geo}} V_{\text{Geo}}^\top
\end{equation}

We extract the top-$r$ singular components that capture the principal geometric information. The low-rank adapters $A_{\text{Geo}}$ and $B_{\text{Geo}}$ are then initialized to approximate this geometry-aware subspace:
% 中文：我们提取能够表征主要几何信息的前 $r$ 个奇异分量；随后将低秩适配器 $A_{\\text{Geo}}$ 与 $B_{\\text{Geo}}$ 初始化，使其近似该几何感知子空间。

\begin{equation}
    A_{\text{Geo}} = \Sigma_{\text{Geo}[:r, :r]}^{1/2} V_{\text{Geo}[:, :r]}^\top
\end{equation}
\begin{equation}
    B_{\text{Geo}} = U_{\text{Geo}[:, :r]} \Sigma_{\text{Geo}[:r, :r]}^{1/2}
\end{equation}
By this design, the initial product $B_{\text{Geo}}A_{\text{Geo}}$ constructs the rank-$r$ approximation of $W_{\text{Geo}}$.
% 中文：通过此设计，初始乘积 $BA$ 构造了 $W_{\text{Geo}}$ 的秩-$r$ 近似。

Crucially, to ensure the model's output remains unchanged at initialization and to preserve core capabilities during training, we follow the standard LoRA scaling and compute a Residual Matrix $W_{\text{res}}$ by subtracting the scaled initialized adapters from the original weights:
% 中文：关键在于，为确保模型输出在初始化时保持不变并在训练中保留核心能力，我们遵循 LoRA 的标准缩放方式，并将 \textbf{残差矩阵} $W_{\text{res}}$ 定义为原始权重减去缩放后的初始化适配器：
\begin{equation}
    W_{\text{res}} = W - \frac{\alpha}{r} B_{\text{Geo}}A_{\text{Geo}}
\end{equation}
During the forward pass, $W_{\text{res}}$ is kept \textbf{frozen}. The hidden state $h$ is computed as:
% 中文：在前向传播中，$W_{\text{res}}$ 保持\textbf{冻结}。隐层状态 $h$ 计算如下：
\begin{equation}
    h = \underbrace{W_{\text{res}}}_{\text{Frozen}} x + \underbrace{\frac{\alpha}{r} B_{\text{Geo}}A_{\text{Geo}}}_{\text{Trainable}} x
\end{equation}
This construction keeps the model function-preserving at initialization (since $W_{\text{res}}x + \frac{\alpha}{r}B_{\text{Geo}}A_{\text{Geo}}x = Wx$) and enforces a hard structural constraint: the optimizer can only update the geometry-aligned manifold parameterized by $A_{\text{Geo}}$ and $B_{\text{Geo}}$, while $W_{\text{res}}$ acts as a stability anchor preventing the erosion of pre-trained representations.
% 中文：该构造保证初始化时函数保持不变（因为 $W_{\text{res}}x + \frac{\alpha}{r}BAx = Wx$），并施加了硬性的结构约束：优化器只能更新由 $A$ 和 $B$ 参数化的几何对齐流形，而 $W_{\text{res}}$ 充当稳定锚点，防止预训练表征被侵蚀。

\subsection{Geometric Prior Construction}
To instantiate a geometry-aware update region, we construct a geometry-constrained matrix $W_{\text{Geo}}$ using a masking strategy. This masking is consistent with prior observations on spectral and magnitude structure and stability in fine-tuning~\cite{lift,1offtheprincipals}.
% 中文：为实例化几何感知的更新区域，我们通过掩码构造几何约束矩阵 $W_{\text{Geo}}$。该思路与既有工作中关于权重谱/幅值结构与微调稳定性的观察一致~\cite{lift,1offtheprincipals}。

\begin{table*}[t]
\centering
\small
\setlength{\tabcolsep}{5pt}
\renewcommand{\arraystretch}{1.08}
\caption{Main results on in-distribution (ID) mathematical RLVR benchmarks and out-of-distribution (OOD) tasks. Base denotes the original model before RLVR. Best results are in bold, and second-best results are underlined.}
\label{tab:main_results}
\resizebox{\textwidth}{!}{
\begin{tabular}{lcccc|ccccc}
\toprule
& \multicolumn{4}{c|}{\textbf{In-Distribution (ID)}} & \multicolumn{5}{c}{\textbf{Out-of-Distribution (OOD)}} \\
\cmidrule(lr){2-5} \cmidrule(lr){6-10}
\textbf{Method} & AIME24 & AIME25 & MATH500 & OlymMATH & HumanEval & GPQA & MMLU & IFEval & TruthfulQA \\
\midrule

\multicolumn{10}{c}{\textit{Qwen3-8B}} \\
\midrule

Base     & 13.33 & 11.67 & 71.20 &  9.75 & 76.83 & 36.91 & 71.94 & \textbf{54.32} & \textbf{68.91} \\
FullFT   & \underline{23.33} & \textbf{22.08} & \textbf{78.40} & 11.25 & 76.83 & 36.91 & 71.94 & 50.45 & 65.65 \\
SparseFT & 22.92 & 21.25 & 76.80 & 11.50 & 79.50 & 37.20 & 74.20 & 50.95 & 66.05 \\
LoRA     & 19.58 & 19.58 & 75.60 & 10.75 & \underline{81.10} & 37.50 & \underline{75.65} & 52.13 & 66.82 \\
PiSSA    & 22.50 & 20.42 & 74.40 & \underline{11.75} & 71.95 & 36.16 & 73.89 & 48.74 & 65.95 \\
MiLoRA   & 20.42 & 19.58 & 76.20 & 11.50 & 78.66 & \textbf{38.26} & 74.51 & 51.85 & 66.46 \\
\rowcolor{blue!5}
GeoRA & \textbf{23.75} & \underline{21.67} & \underline{78.00} & \textbf{12.75} & \textbf{82.93} & \underline{37.92} & \textbf{75.96} & \underline{53.73} & \underline{68.85} \\
\midrule

\multicolumn{10}{c}{\textit{Llama-3.1-8B}} \\
\midrule

Base     &  9.58 &  2.08 & 51.00 &  3.25 & 65.20 & 31.15 & 68.40 & \textbf{79.99} & \textbf{62.71} \\
FullFT   & \underline{18.33} & \underline{8.25} & \textbf{62.40} & \underline{8.50} & 65.20 & 31.15 & 68.40 & 75.92 & 59.40 \\
SparseFT & 17.92 &  8.10 & 61.50 &  8.25 & 67.80 & 31.65 & 69.10 & 76.61 & 60.19 \\
LoRA     & 15.42 &  6.25 & 58.20 &  6.75 & \underline{69.80} & \underline{32.10} & 69.80 & 77.93 & \underline{61.76} \\
PiSSA    & 17.50 &  7.92 & 60.50 &  7.75 & 67.50 & 31.80 & 69.20 & 74.83 & 59.83 \\
MiLoRA   & 16.25 &  7.08 & 59.10 &  7.25 & 68.20 & 32.00 & \underline{70.50} & 78.31 & 61.47 \\
\rowcolor{blue!5}
GeoRA & \textbf{18.54} & \textbf{8.75} & \underline{61.90} & \textbf{8.85} & \textbf{70.80} & \textbf{32.65} & \textbf{70.95} & \underline{78.72} & 61.61 \\
\bottomrule
\end{tabular}
}
\end{table*}

\begin{table*}[t]
\centering
\small
\setlength{\tabcolsep}{5pt}
\renewcommand{\arraystretch}{1.08}
\caption{Results on additional medical and coding RLVR tasks. Best results are in bold, and second-best results are underlined.}
\label{tab:cross_domain_rlvr}
\resizebox{\textwidth}{!}{
\begin{tabular}{lcccc|cccc}
\toprule
& \multicolumn{4}{c|}{\textbf{Medical}} & \multicolumn{4}{c}{\textbf{Coding}} \\
\cmidrule(lr){2-5} \cmidrule(lr){6-9}
\textbf{Method} & MedQA & MedMCQA & PubMedQA & Average & LiveCodeBench & HumanEval & MBPP & Average \\
\midrule
Base   & 58.03 & 56.42 & 74.94 & 63.13 & 65.75 & 87.66 & 79.40 & 77.60 \\
FullFT & \underline{75.32} & \textbf{64.56} & \underline{80.12} & \underline{73.33} & \textbf{67.75} & \textbf{90.26} & \underline{81.40} & \textbf{79.80} \\
LoRA   & 74.23 & 62.12 & 79.54 & 71.96 & 67.25 & 88.96 & 81.00 & 79.07 \\
\rowcolor{blue!5}
GeoRA  & \textbf{76.12} & \underline{64.31} & \textbf{80.64} & \textbf{73.69} & \textbf{67.75} & \underline{89.61} & \textbf{81.60} & \underline{79.65} \\
\bottomrule
\end{tabular}
}
\end{table*}

The Spectral Prior ($M_{\text{Spec}}$) promotes geometric stability by selecting the bottom $\rho$-fraction of entries from the rank-$r$ approximation $\hat{W}_r$. The mask is defined as:
\begin{equation}
    (M_{\text{Spec}})_{i,j} = \mathbb{I}\left( |(\hat{W}_r)_{i,j}| \leq \tau_{\text{Spec}}(\rho) \right)
\end{equation}
where $\tau_{\text{Spec}}(\rho)$ is the $\rho$-th quantile of the absolute values in $\hat{W}_r$. Intuitively, this mask suppresses high-magnitude (and typically high-curvature) components and constrains updates to a more stable, low-magnitude region, improving spectral stability under RLVR.
% 中文：谱先验 ($M_{\text{Spec}}$) 在秩-$r$ 近似矩阵 $\hat{W}_r$ 上保留幅值处于底部 $\rho$ 比例的元素；其中 $\tau_{\text{Spec}}(\rho)$ 表示 $|\hat{W}_r|$ 的第 $\rho$ 分位数（quantile）。该掩码通过抑制主方向的高幅/高曲率成分，将更新约束在结构上更稳定的低幅区域，从而提升 RLVR 训练过程中的谱稳定性。
Similarly, the Euclidean Prior ($M_{\text{Euc}}$) selects low-magnitude weights to capture parameter plasticity, using the same sparsity ratio $\rho$:
\begin{equation}
    (M_{\text{Euc}})_{i,j} = \mathbb{I}\left( |W_{i,j}| \leq \tau_{\text{Euc}}(\rho) \right)
\end{equation}
Here, $\tau_{\text{Euc}}(\rho)$ represents the $\rho$-th quantile of $|W|$. The final geometry-constrained matrix $W_{\text{Geo}}$ is formed by the \textbf{union} of these two stable subspaces:
\begin{equation}
    W_{\text{Geo}} = W \odot (M_{\text{Spec}} \cup M_{\text{Euc}})
\end{equation}
This union ensures that the optimized weights retain the flexibility of small parameters while respecting the spectral constraints of the pre-trained model.
% 欧氏先验 ($M_{\text{Euc}}$) 同样基于比例 $\rho$ 选择原始权重中的低幅值部分，以捕捉参数的可塑性：此处 $\tau_{\text{Euc}}(\rho)$ 对应 $|W|$ 的第 $\rho$ 分位数。最终的几何约束矩阵 $W_{\text{Geo}}$ 由这两个稳定子空间的并集构成， 这种并集策略确保了模型既能保留微小权重的可塑性，又能通过谱约束避免破坏预训练模型的特征结构。

\section{Experiments}

\subsection{Experimental Setup}

We compare GeoRA against MiLoRA~\cite{milora}, PiSSA~\cite{pissa}, LoRA~\cite{lora}, Sparse Fine-Tuning (SparseFT)~\cite{1offtheprincipals}, and Full Fine-Tuning (FullFT). Our main experiments are conducted on mathematical RLVR, where we fine-tune Qwen3-8B-Base~\cite{qwen3} and Llama-3.1-8B-Instruct~\cite{llama3} on the DeepMath-103K dataset~\cite{deepmath} using the GRPO algorithm~\cite{grpo}, with a fixed rank $r=16$ and sparsity ratio $\rho=0.2$. We include both base and instruction-tuned backbones to evaluate robustness across model variants. In addition, we further validate our method on mathematical RLVR tasks with the 4B and 1.5B models on the GSM8K~\cite{gsm8k} dataset; details are provided in the Appendix~\ref{app:exp_details}.

For in-distribution (ID) evaluation, we assess mathematical reasoning performance on AIME24, AIME25~\cite{aime2024}, MATH-500~\cite{math500}, and OlymMATH~\cite{olymmath}. For out-of-distribution (OOD) evaluation, we consider HumanEval (Coding)~\cite{humaneval}, GPQA (Science)~\cite{gpqa}, MMLU (General Knowledge)~\cite{mmlu}, IFEval (Instruction Following)~\cite{ifeval}, and TruthfulQA (Truthfulness)~\cite{truthfulqa} to measure cross-domain generalization and capability preservation. We also report the pre-RLVR Base model performance for direct comparison.

Beyond the main mathematical setting, we further evaluate GeoRA on additional RLVR domains, including medical reasoning and coding. For medical RLVR, we train Llama-3.1-8B-Instruct on AlphaMed19K~\cite{alphamed19k} and report results on MedQA~\cite{medqa}, MedMCQA~\cite{medmcqa}, and PubMedQA~\cite{pubmedqa}. For coding, we train Qwen3-32B on Eurus-2-RL-Data~\cite{eurus2} and evaluate on LiveCodeBench~\cite{livecodebench}, HumanEval, and MBPP~\cite{mbpp}. The implementation details and hyperparameter settings are provided in the Appendix~\ref{app:exp_details}.
% 中文：除了主要的数学场景，我们在医疗推理和代码等其他 RLVR 领域对 GeoRA 进行了进一步评估。对于医疗 RLVR，我们在 AlphaMed19K 数据集上训练 Llama-3.1-8B-Instruct 模型并汇报 MedQA、MedMCQA 和 PubMedQA 的结果。对于代码 RLVR，我们在 Eurus-2 数据集上训练 Qwen3-32B 模型并在 LiveCodeBench、HumanEval 和 MBPP 上进行评估。所有实现细节和超参数设置均在附录中提供。

\subsection{Main Results}

\textbf{Mathematical Reasoning Performance.}
Table~\ref{tab:main_results} shows that GeoRA achieves the strongest overall performance on mathematical RLVR benchmarks across both backbones. It obtains the best results on most AIME and OlymMATH settings, while remaining competitive on MATH500. In particular, GeoRA consistently outperforms LoRA, MiLoRA, and PiSSA on challenging competition-style benchmarks, suggesting that geometry-aware initialization provides a more suitable inductive bias for RLVR than classic low-rank initialization.

\textbf{Out-of-Distribution Performance.}
GeoRA also generalizes favorably to OOD tasks. Across both backbones, it achieves the strongest or near-strongest results on HumanEval, GPQA, and MMLU, while generally retaining higher capability than other fine-tuning baselines on IFEval and TruthfulQA. These results indicate that constraining updates to geometry-aligned subspaces can improve in-domain reasoning performance while reducing interference with pre-existing general capabilities.

\textbf{Extension Beyond Mathematical RLVR.}
Beyond the main mathematical RLVR setting, we further evaluate GeoRA on medical and coding RLVR tasks. As shown in Table~\ref{tab:cross_domain_rlvr}, GeoRA consistently outperforms LoRA across both domains and remains competitive with FullFT, suggesting that its advantage is not limited to mathematical reasoning.

\begin{figure}[t]
  \centering
  \includegraphics[width=0.95\columnwidth]{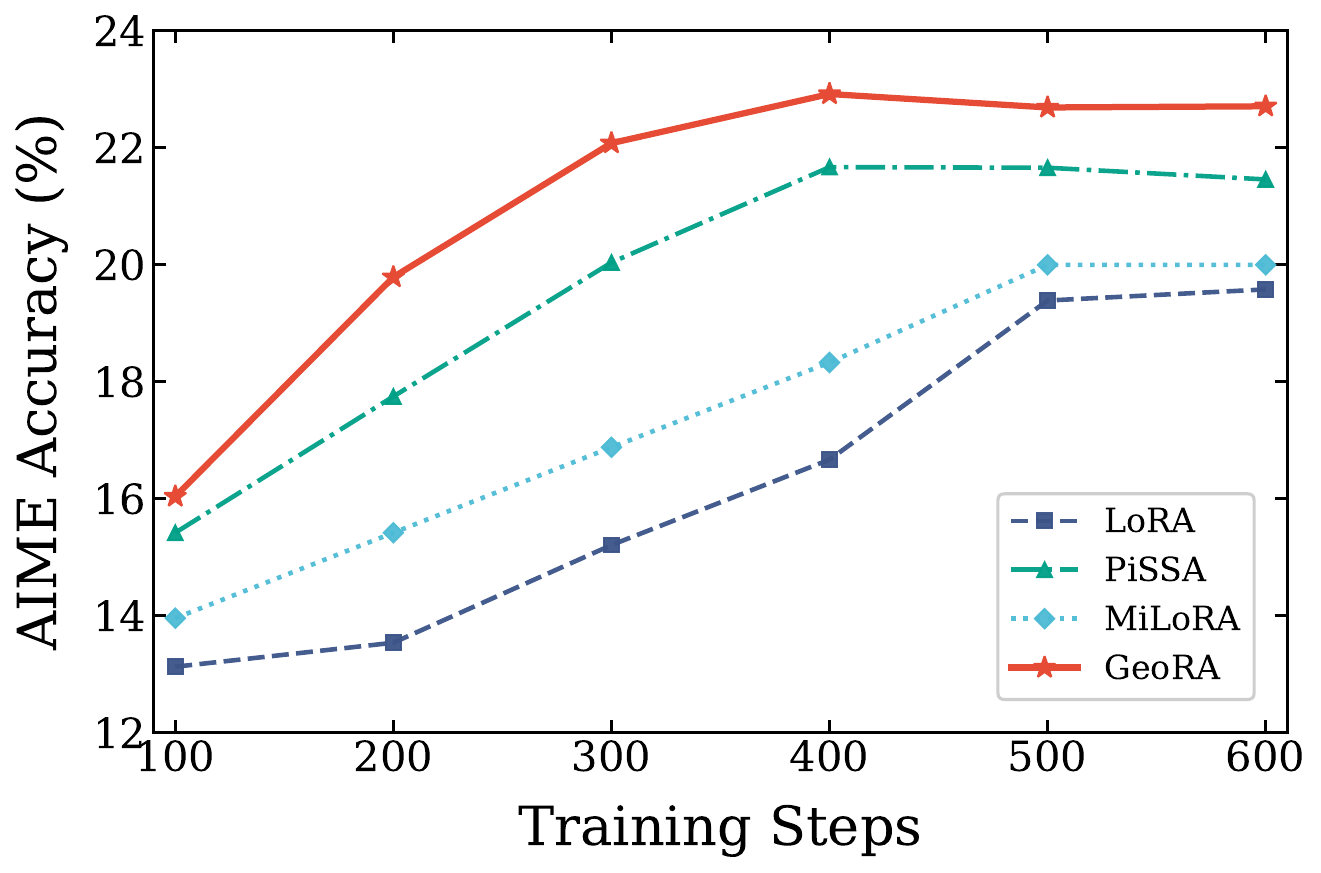}
\caption{Training dynamics of Qwen3-8B as evaluated on the AIME benchmark (average of 2024 and 2025). GeoRA remains consistently top-performing throughout training.}
  \label{fig:aime}
\end{figure}

\subsection{Training Stability and Efficiency}
\textbf{Training Dynamics.}
Figure~\ref{fig:aime} shows the training dynamics of Qwen3-8B as evaluated on the AIME benchmark. GeoRA remains consistently top-performing throughout training and reaches strong performance substantially earlier than other low-rank baselines. In contrast, LoRA, MiLoRA, and PiSSA improve more slowly and plateau at lower levels. This indicates that GeoRA not only yields better final accuracy, but also provides a more favorable optimization trajectory under RLVR.

\begin{figure}[t]
    \centering
    \includegraphics[width=\columnwidth]{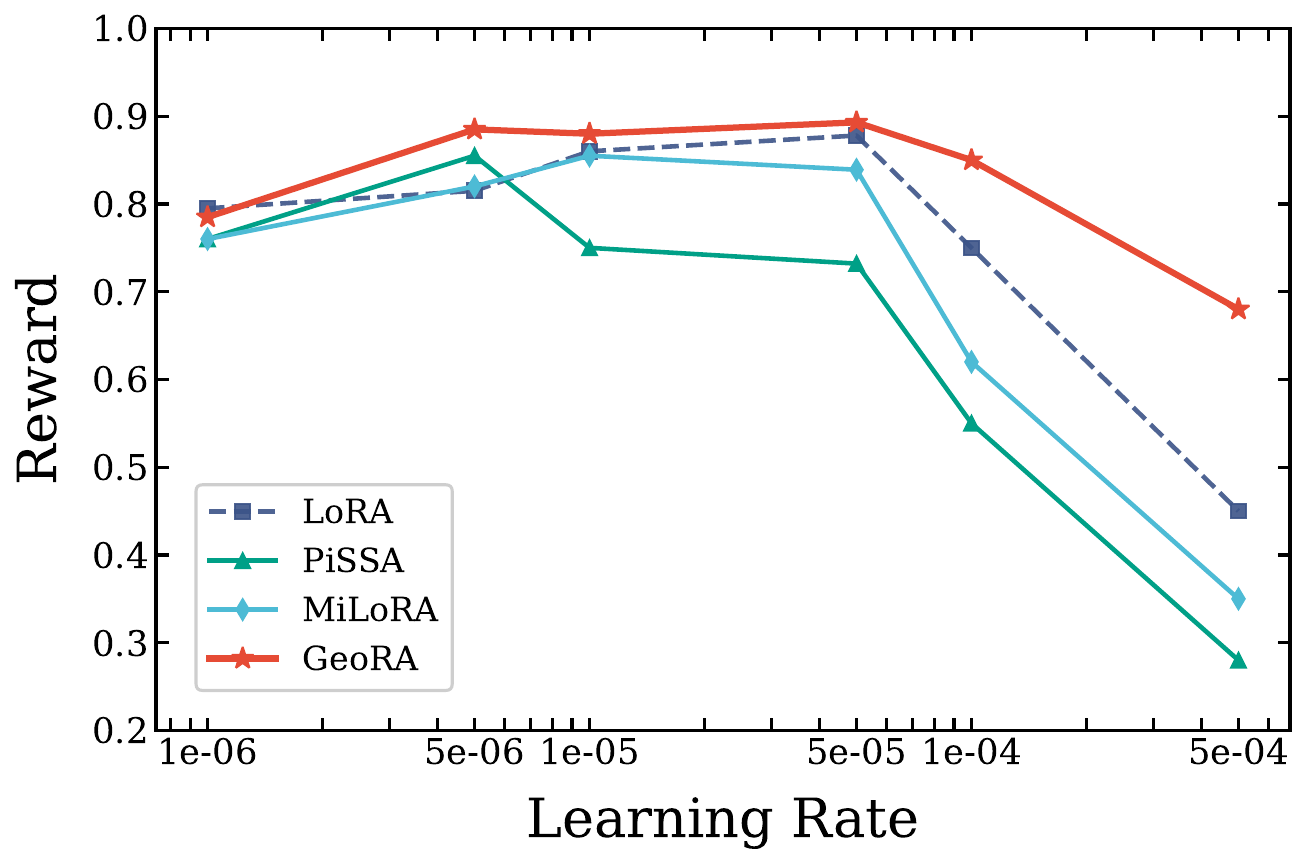}
    \caption{Performance comparison across different learning rates. GeoRA demonstrates superior stability and robust convergence even at higher learning rates.}
    \label{fig:lr_analysis}
\end{figure}

\textbf{Hyperparameter Robustness.}
We first evaluate performance under different learning rates on Qwen3-4B in Figure~\ref{fig:lr_analysis}. GeoRA maintains high reward across a broad range of learning rates, while other low-rank baselines are much more sensitive to this choice. In particular, PiSSA and MiLoRA degrade rapidly under larger learning rates, and LoRA also exhibits a clear drop in performance at the high end of the sweep. These results suggest that GeoRA is more robust to hyperparameter variation and requires less delicate tuning in practice. Similar robustness trends are also observed for other hyperparameters, including rank and sparsity, as shown in Appendix~\ref{app:robustness}.

\begin{figure*}
    \centering
    % 左图：Reward
    \begin{subfigure}{0.49\linewidth}
        \centering
        \includegraphics[width=\linewidth]{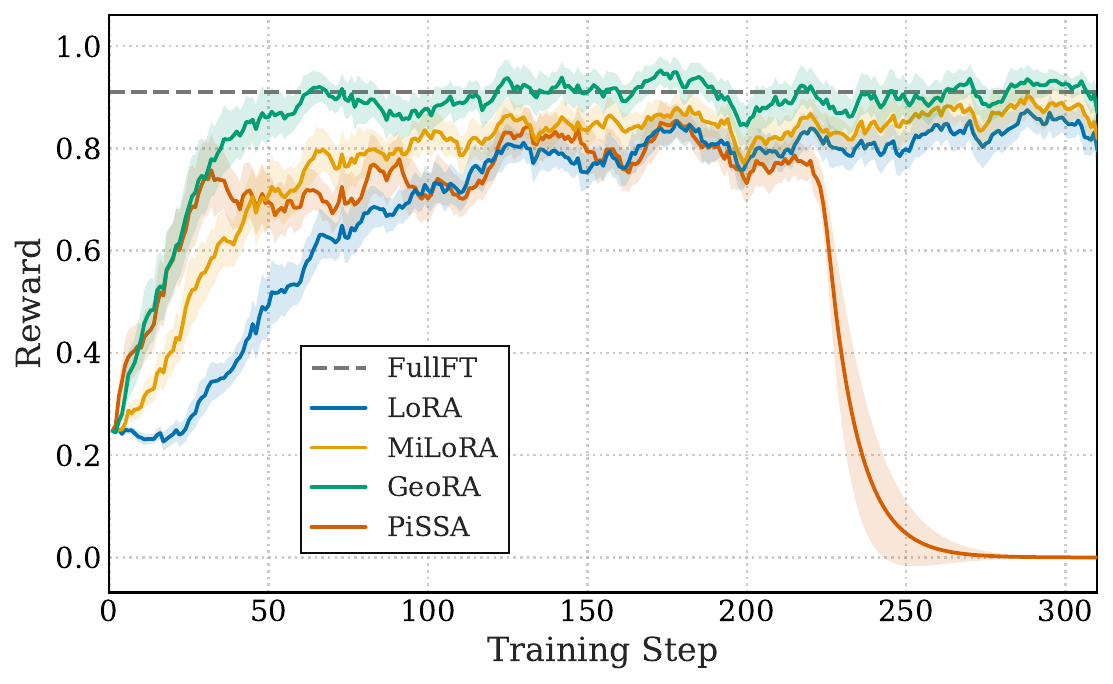}
        \caption{Reward Trajectories}
        \label{fig:robustness_reward}
    \end{subfigure}
    \hfill
    % 右图：KL
    \begin{subfigure}{0.49\linewidth}
        \centering
        \includegraphics[width=\linewidth]{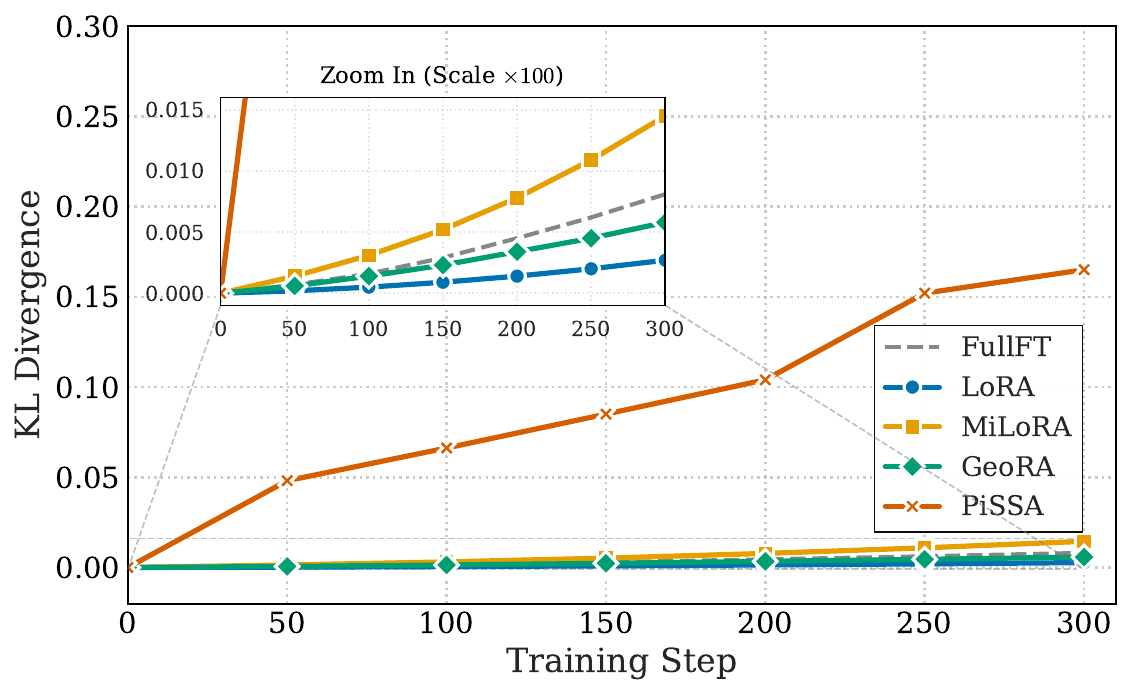}
        \caption{KL Divergence}
        \label{fig:robustness_kl}
    \end{subfigure}

    \caption{
  Training Stability and Constraint Adherence. Results on Qwen3-4B show that GeoRA demonstrates superior robustness under aggressive learning rates ($5 \times 10^{-5}$).
  }
    \label{fig:robustness_analysis}
\end{figure*}

\textbf{Stability Under Aggressive Optimization.}
To stress-test optimization stability, we further compare reward and KL divergence under an aggressive learning rate in Figure~\ref{fig:robustness_analysis}. GeoRA maintains the highest reward trajectory without collapse, whereas PiSSA suffers a catastrophic drop late in training. At the same time, GeoRA keeps KL divergence at a low and controlled level throughout training. These results are highly relevant to RLVR optimization, which often requires stable policy updates within trust-region-like boundaries. The combination of higher reward and smoother KL behavior suggests that GeoRA improves exploration and policy refinement without inducing the destabilizing updates caused by geometry-misaligned adaptation.

\textbf{Efficiency.}
Table~\ref{tab:efficiency_compact} compares the training efficiency of FullFT, SparseFT, and GeoRA. GeoRA updates only a tiny fraction of parameters, reducing trainable parameters by 99.5\% relative to FullFT, while also lowering VRAM usage and improving per-iteration training speed. Compared with SparseFT, GeoRA avoids the overhead of unstructured sparse computation and therefore translates its parameter efficiency into actual hardware efficiency. Although GeoRA introduces an SVD-based initialization step, this cost is a one-time preprocessing overhead and is negligible compared with RLVR training; detailed profiling is provided in the appendix. Overall, these results show that GeoRA improves not only effectiveness and stability, but also practical training efficiency.

\begin{table}[h]
    \centering
    \caption{Efficiency Comparison between Full FT, SparseFT, and GeoRA. Relative reductions compared to Full FT are shown in parentheses. }
    \label{tab:efficiency_compact}
    \setlength{\tabcolsep}{2pt} % 稍微增加一点间距以容纳百分号
    \resizebox{0.8\columnwidth}{!}{
    \begin{tabular}{l c c c}
    \toprule
    \textbf{Method} & Params (B) & Time (s/it) & VRAM (\%) \\
    \midrule
    Full FT
     & 8.00 
     & 231 
     & 95.73 \\
    \addlinespace[4pt] % 增加行间距以区分不同方法
    SparseFT
     & \begin{tabular}{@{}c@{}} 2.56 \\ (-68.0\%) \end{tabular} 
     & \begin{tabular}{@{}c@{}} 256 \\ (+10.8\%) \end{tabular} 
     & \begin{tabular}{@{}c@{}} 81.25 \\ (-15.1\%) \end{tabular} \\
    \addlinespace[4pt]
    GeoRA
     & \begin{tabular}{@{}c@{}} \textbf{0.04} \\ (-99.5\%) \end{tabular} 
     & \begin{tabular}{@{}c@{}} \textbf{185} \\ (-19.9\%) \end{tabular} 
     & \begin{tabular}{@{}c@{}} \textbf{68.43} \\ (-28.5\%) \end{tabular} \\
    \bottomrule
    \end{tabular}
    }
    \end{table}

\begin{table*}[t]
    \centering
    \resizebox{0.8\textwidth}{!}{
    \begin{tabular}{l|c|cccc|c}
        \toprule
        \textbf{Method} & Reward & AIME24 & AIME25 & MATH-500 & OlymMATH & Average \\
        \midrule
        \hspace{3mm}GeoRA & \textbf{0.88} & \textbf{13.33} & \textbf{9.17} & \textbf{73.40} & \textbf{5.75} & \textbf{25.41} \\
        \midrule
        \multicolumn{7}{l}{\textit{Ablation on Initialization Strategy}} \\
        \hspace{3mm} Random-$r$ Init & 0.85 & 12.50 & 8.50 & 72.10 & 5.25 & 24.60 \\
        \hspace{3mm} Tail-$r$ Init & 0.82 & 11.67 & 7.50 & 70.80 & 4.50 & 23.40 \\
        \midrule
        \multicolumn{7}{l}{\textit{Ablation on Geometric Masks}} \\
        \hspace{3mm} w/o $M_{\text{Spec}}$ & 0.86 & 12.50 & 8.33 & 72.00 & 4.75 & 24.40 \\
        \hspace{3mm} w/o $M_{\text{Euc}}$ & 0.83 & 13.33 & 8.75 & 72.80 & 5.50 & 25.10 \\
        \bottomrule
    \end{tabular}
    }
\caption{Ablation study on Qwen3-4B with different initialization strategies and geometric mask variants.}
    \label{tab:ablation}
\end{table*}

\subsection{Ablation and Analysis}
\textbf{Impact of Initialization Strategy.}
Table~\ref{tab:ablation} shows that GeoRA’s geometry-aware initialization is critical to its performance. Replacing the proposed initialization with Random-\(r\) initialization leads to a clear drop across all benchmarks, while Tail-\(r\) initialization performs even worse. This indicates that the benefit of GeoRA does not come from low-rank adaptation alone, but from aligning the initialization with the effective RLVR update subspace. In particular, the poor performance of Tail-\(r\) suggests that simply selecting low-energy directions is insufficient; the trainable directions must still capture the dominant structure within the geometry-constrained subspace.

\textbf{Impact of Geometric Masks.}
We further ablate the two mask components used to construct the geometry-constrained subspace. Removing either \(M_{\text{Spec}}\) or \(M_{\text{Euc}}\) consistently degrades performance, confirming that both are necessary for strong results. This suggests that the two priors play complementary roles: the spectral mask helps suppress unstable high-energy directions, while the Euclidean mask preserves adaptation flexibility in low-magnitude regions. Their combination therefore provides a better approximation to the effective RLVR update manifold than either component alone.

\textbf{Overall Analysis.}
Taken together, the ablation results show that GeoRA’s gains arise from the full geometry-aware design rather than from any single isolated choice. The geometry-constrained subspace, structured initialization, and frozen residual anchor work together to improve optimization quality while preserving pre-trained structure. This is consistent with the stability and efficiency results in the previous section, and further supports our claim that RLVR benefits from a low-rank parameterization explicitly aligned with its update geometry.

\section{Mechanism Analysis}

\subsection{Geometric Priors}
GeoRA constructs its update subspace from two geometric priors: a spectral prior and a Euclidean prior. The motivation is that RLVR updates are highly selective, tending to exploit under-utilized yet plastic regions of the pre-trained model while avoiding large modifications to dominant principal directions. Accordingly, $M_{\text{Spec}}$ suppresses high-energy components in the principal subspace, whereas $M_{\text{Euc}}$ selects low-magnitude parameters in the original weight space. Their union defines a subspace that is both stable and expressive for RLVR adaptation.
\begin{table}[t]
\centering
\small
\setlength{\tabcolsep}{4.5pt}
\renewcommand{\arraystretch}{1.55}
\caption{Overlap analysis of the spectral prior $M_{\text{Spec}}$ and Euclidean prior $M_{\text{Euc}}$ on Qwen3-8B with $\rho=0.2$.}
\label{tab:mask_overlap}
\begin{tabular}{lcccc}
\toprule
\textbf{Group} & $M_{\text{Spec}}$ & $M_{\text{Euc}}$ & Intersection & Jaccard \\
\midrule
MLP  & 20.0\% & 20.0\% & 4.63\% & 0.131 \\
Attn & 20.0\% & 20.0\% & 4.25\% & 0.119 \\
\textbf{All} & \textbf{20.0\%} & \textbf{20.0\%} & \textbf{4.55\%} & \textbf{0.128} \\
\bottomrule
\end{tabular}
\end{table}

We further verify that these two priors are complementary rather than redundant in Table~\ref{tab:mask_overlap}. On Qwen3-8B with $\rho=0.2$, both $M_{\text{Spec}}$ and $M_{\text{Euc}}$ select 20.0\% of parameters, but their overlap remains small: the overall intersection is only 4.55\%, with a Jaccard index of 0.128. Similar patterns are observed in both MLP layers (4.63\%, 0.131) and attention layers (4.25\%, 0.119). This limited overlap indicates that the two masks capture largely distinct parameter subsets and therefore serve complementary geometric roles. This is also consistent with the ablation results in Table~\ref{tab:ablation}: removing either prior leads to a clear performance drop.

\subsection{Low-Rank Structure}

\begin{figure}[t]
    \centering
    \begin{subfigure}[t]{\linewidth}
        \centering
        \includegraphics[width=\linewidth]{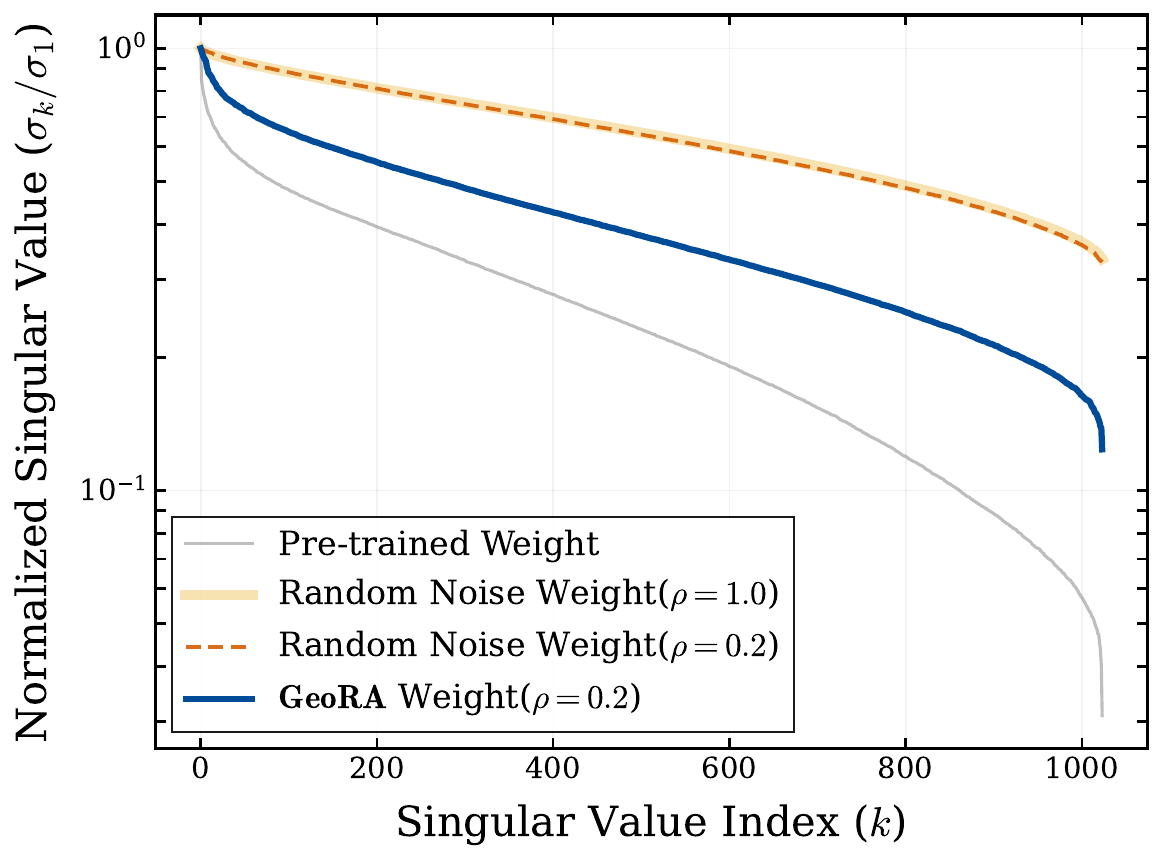}
        \caption{Singular value spectrum of the geometry-constrained subspace $W_{\text{Geo}}$.}
        \label{fig:spectrum_geo}
    \end{subfigure}
    
    \vspace{0.5em}
    
    \begin{subfigure}[t]{\linewidth}
        \centering
        \includegraphics[width=\linewidth]{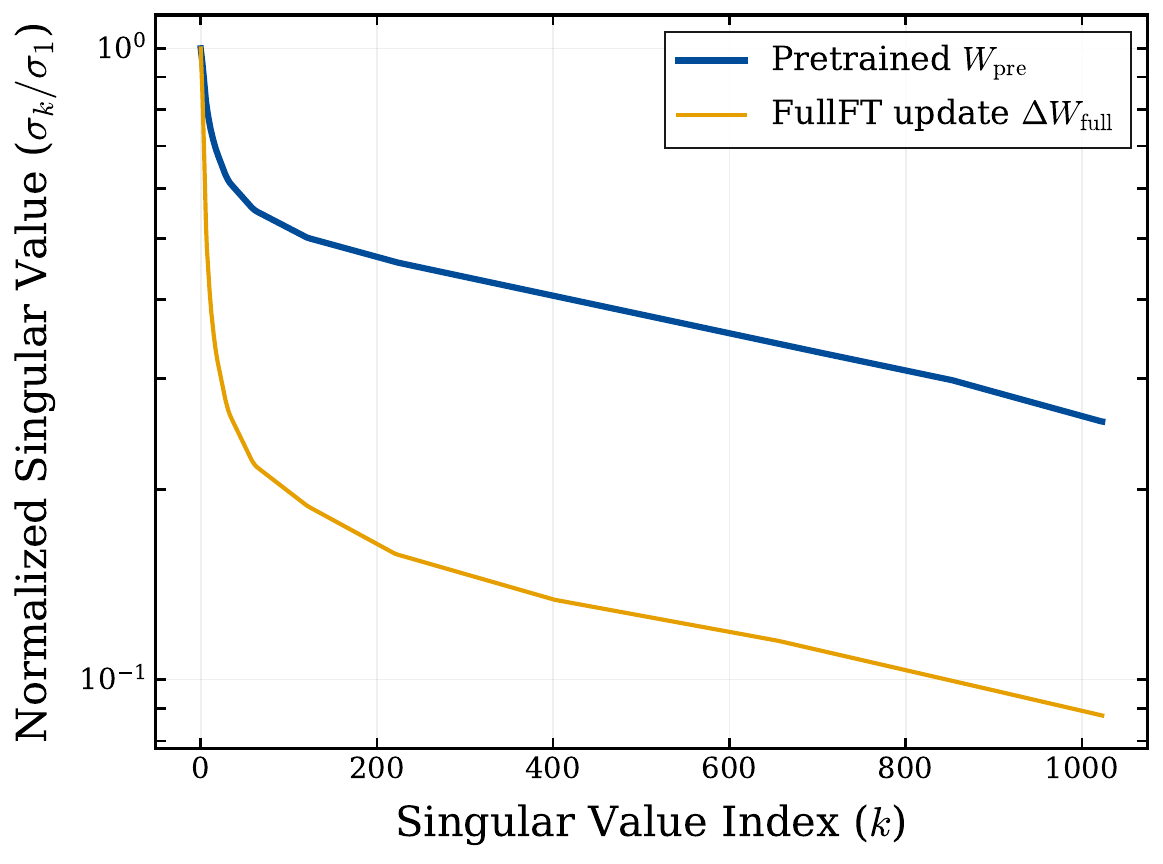}
        \caption{Singular value spectrum of the full fine-tuning update $\Delta W_{\text{FullFT}}$.}
        \label{fig:spectrum_fullft}
    \end{subfigure}
    \caption{Spectral analysis of RLVR update structure.}
    \label{fig:spectrum}
\end{figure}

To validate the structural basis of GeoRA, we analyze the singular value spectrum of the geometry-constrained subspace. We compare $W_{\text{Geo}}$ with the original pre-trained weights, as well as random noise matrices at both full and sparse density. Figure~\ref{fig:spectrum_geo} yields two key insights. First, sparsity does not by itself induce low-rankness. The spectrum of sparse random noise is nearly indistinguishable from that of dense random noise, and both follow a relatively flat decay pattern. This indicates that unstructured sparsity alone remains essentially isotropic in the spectral domain. Second, $W_{\text{Geo}}$ preserves a pronounced heavy-tailed spectrum similar to that of the pre-trained weights, with most spectral mass concentrated in a small number of leading components. This shows that the selected update region inherits a structured and compressible low-rank form rather than behaving like random sparse noise.

We further find that the actual FullFT update, $\Delta W_{\text{FullFT}} = W_{\text{FullFT}} - W_{\text{Pretrain}}$, exhibits a similarly compressible heavy-tailed spectrum. This strongly supports the low-rank inductive bias of GeoRA: the effective RLVR update space is not isotropic, but already structured and highly compressible. GeoRA therefore captures an intrinsic property of RLVR updates, instead of imposing an artificial low-rank constraint.

\begin{table}[t]
\centering
\small
\setlength{\tabcolsep}{3pt}
\renewcommand{\arraystretch}{1.45}
\caption{Geometric mechanism analysis.}
\label{tab:mechanism_combined}
\begin{tabular}{lcccccc}
\toprule
& \multicolumn{3}{c}{\textbf{Llama-3.1-8B}} & \multicolumn{3}{c}{\textbf{Qwen3-8B}} \\
\cmidrule(lr){2-4} \cmidrule(lr){5-7}
\textbf{Method} & \textbf{NSS}$\downarrow$ & $\mathcal{S}_H\downarrow$ & $\mathcal{S}_T\uparrow$ & \textbf{NSS}$\downarrow$ & $\mathcal{S}_H\downarrow$ & $\mathcal{S}_T\uparrow$ \\
\midrule
PiSSA  & 0.395 & 0.98  & 0.01 & 0.418 & 0.95  & 0.03 \\
LoRA   & 0.214 & 0.15  & 0.15 & 0.235 & 0.18  & 0.18 \\
MiLoRA & 0.125 & 0.12  & 0.92 & 0.132 & 0.16  & 0.90 \\
\textbf{GeoRA} & \textbf{0.092} & \textbf{0.005} & \textbf{0.98} & \textbf{0.096} & \textbf{0.015} & \textbf{0.96} \\
\bottomrule
\end{tabular}
\end{table}

\subsection{Spectral Efficiency and Alignment}
Building on the low-rank structure of RL updates, we validate GeoRA's geometric superiority via Normalized Spectral Shift (NSS) and Subspace Alignment(Table~\ref{tab:mechanism_combined}). First, NSS quantifies the topological distortion of the pre-trained manifold:
\begin{equation}
    \small
    \text{NSS} = \frac{\|\sigma(W^{\text{tuned}}) - \sigma(W)\|_2}{\|\sigma(W)\|_2}.
    \label{eq:nss}
\end{equation}
While PiSSA exhibits high NSS ($>0.39$), indicating aggressive structural modification, GeoRA maintains minimal distortion ($\approx 0.09$), confirming it maximizes reward acquisition while preserving fundamental features.

To pinpoint the update locus, we calculate the alignment between $\Delta W$ and the pre-trained singular vectors $V$:
\begin{equation}
    \small
    \mathcal{S}(k) = \frac{\| \Delta W v_k \|_2}{\| \Delta W \|_F} \approx |\cos(\theta_k)|.
    \label{eq:alignment}
\end{equation}
GeoRA shows a distinct signature: it avoids the high-energy principal subspace ($\mathcal{S}_{\text{Head}} \le 0.02$)---enhancing stability---and efficiently adapts the geometry-constrained tail ($\mathcal{S}_{\text{Tail}} \ge 0.96$). In contrast, PiSSA's instability is explained by its high overlap with the head subspace ($\approx 0.98$).

\section{Conclusion}
In this paper, we bridge the gap between parameter-efficient fine-tuning and the distinctive optimization geometry of RLVR. Our study suggests that existing PEFT methods often yield suboptimal results in RLVR due to a structural mismatch. SFT-oriented low-rank methods prioritize directions that may be misaligned with the geometry-constrained updates favored by RLVR, while some sparse update methods better reflect RLVR dynamics but fail to translate into practical hardware efficiency. GeoRA addresses these challenges by identifying a compressible, geometry-aligned update subspace and parameterizing it through structured SVD initialization with a frozen residual anchor. Experiments on Qwen and Llama models from 1.5B to 32B parameters show that GeoRA consistently improves mathematical RLVR performance, extends effectively to medical and coding RLVR settings, and exhibits stronger generalization with less forgetting on out-of-domain tasks. Overall, GeoRA provides an effective, stable, and hardware-efficient adaptation strategy for RLVR.

\section*{Limitations}

Despite its effectiveness, GeoRA has certain limitations. First, the initialization process requires performing a truncated SVD and dual-masking operations. While these represent a one-time computational cost at the beginning of training, they introduce an additional pre-processing step compared to the random initialization used in standard LoRA. Second, our experiments primarily focus on RLVR in reasoning domains. Although GeoRA demonstrates strong performance on these tasks, its generalizability needs to be further validated across a broader range of model architectures and diverse reinforcement learning scenarios beyond verifiable rewards.

\bibliography{custom}
\clearpage
\appendix

\section{Algorithm}
\label{app:algorithm}
\begin{algorithm}
\caption{GeoRA Initialization Algorithm}
\label{alg:geora_init}
\begin{algorithmic}[1]
\REQUIRE Pre-trained weights $W \in \mathbb{R}^{m \times n}$, rank $r$, sparsity ratio $\rho$.
\ENSURE Initialized adapters $A, B$, Frozen residual matrix $W_{\text{res}}$.

\STATE \textbf{Phase 1: Geometric Prior Construction}
\STATE $\hat{W}_r \leftarrow \text{SVD}_r(W)$ \COMMENT{Compute rank-$r$ approximation}
\STATE Determine thresholds $\tau_{\text{spec}}$ and $\tau_{\text{euc}}$ corresponding to bottom $\rho$ quantile
\STATE $M_{\text{spec}} \leftarrow \mathbb{I}(|\hat{W}_r| \le \tau_{\text{spec}})$ \COMMENT{Spectral Prior}
\STATE $M_{\text{euc}} \leftarrow \mathbb{I}(|W| \le \tau_{\text{euc}})$ \COMMENT{Euclidean Prior}
\STATE $W_{\text{geo}} \leftarrow W \odot (M_{\text{spec}} \cup M_{\text{euc}})$ \COMMENT{Union of priors (unconstrained sparsity)}

\STATE \textbf{Phase 2: Geometry-Aware Decomposition}
\STATE $U_g, \Sigma_g, V_g^\top \leftarrow \text{SVD}(W_{\text{geo}})$
\STATE $A \leftarrow U_g[:,:r] \Sigma_g^{1/2}$
\STATE $B \leftarrow \Sigma_g^{1/2} V_g^\top[:,:r]$

\STATE \textbf{Phase 3: Residual Leash Instantiation}
\STATE $W_{\text{res}} \leftarrow W - A B$
\STATE \textbf{Freeze} $W_{\text{res}}$ during training

\RETURN $A, B, W_{\text{res}}$
\end{algorithmic}
\end{algorithm}

\section{GeoRA Initialization Properties}
\label{app:theory}
This section provides lightweight theoretical justifications for GeoRA's initialization and re-parameterization.
% 中文：本节给出 GeoRA 初始化与重参数化的若干轻量级理论说明（用于增强可解释性与“看起来硬核”）。

\subsection{Optimality of the Geo-SVD Initialization}
Proposition 1 (Eckart--Young optimality within $W_{\text{geo}}$).
Let $W_{\text{geo}} \in \mathbb{R}^{m \times n}$ admit an SVD $W_{\text{geo}} = U \Sigma V^\top$, and define the truncated reconstruction $W_{\text{geo}}^{(r)} \triangleq U_{[:, :r]} \Sigma_{[:r,:r]} V_{[:, :r]}^\top$.
Then $W_{\text{geo}}^{(r)}$ is the best rank-$r$ approximation under the Frobenius norm:
\begin{equation}
    W_{\text{geo}}^{(r)} \in \arg\min_{\mathrm{rank}(X)\le r}\ \|W_{\text{geo}} - X\|_F.
\end{equation}
% 中文：命题 1：在 Frobenius 范数下，$W_{\text{geo}}$ 的前 $r$ 个奇异分量给出最优的秩-$r$ 近似（Eckart--Young 定理）。

With the initialization $B \leftarrow U_{[:, :r]} \Sigma_{[:r,:r]}^{1/2}$ and $A \leftarrow \Sigma_{[:r,:r]}^{1/2} V_{[:, :r]}^\top$, we have $BA = W_{\text{geo}}^{(r)}$.
% 中文：用 $B=U\Sigma^{1/2}$、$A=\Sigma^{1/2}V^\top$ 初始化可保证 $BA$ 恰为 $W_{\text{geo}}$ 的最优秩-$r$ 重构。
\subsection{Initialization and Residual Leash}
Proposition 2 (Function-preserving at initialization).
Define $W_{\mathrm{res}} \triangleq W - \frac{\alpha}{r}BA$ and the effective weight $W_{\mathrm{eff}}(A,B) \triangleq W_{\mathrm{res}} + \frac{\alpha}{r}BA$.
Then for any input $x$, the forward pass satisfies $W_{\mathrm{eff}}(A,B)x = Wx$ at initialization, hence GeoRA is function-preserving at $t=0$.
% 中文：命题 2：用 $W_{\text{res}} = W - \frac{\alpha}{r}BA$ 定义残差牵绳后，初始化时 $W_{\text{res}}x + \frac{\alpha}{r}BAx = Wx$，因此函数保持不变。

Gradient mapping (useful for understanding stability).
Let $G \triangleq \frac{\partial \mathcal{L}}{\partial W_{\mathrm{eff}}}$ denote the gradient w.r.t. the effective weight.
By matrix calculus,
\begin{equation}
    \frac{\partial \mathcal{L}}{\partial A} = \frac{\alpha}{r} B^\top G,\quad
    \frac{\partial \mathcal{L}}{\partial B} = \frac{\alpha}{r} G A^\top.
\end{equation}
Freezing $W_{\mathrm{res}}$ constrains learning to the rank-$r$ manifold $\{ \frac{\alpha}{r}BA : A\in\mathbb{R}^{r\times n}, B\in\mathbb{R}^{m\times r}\}$, while the residual leash keeps the model anchored near $W$.
% 中文：梯度映射：冻结 $W_{\text{res}}$ 时，优化仅在秩-$r$ 流形上进行；$W_{\text{res}}$ 作为锚点抑制偏离预训练几何的漂移。

\begin{table}[t]
\centering
\footnotesize
\setlength{\tabcolsep}{3.5pt}
\renewcommand{\arraystretch}{1.45}
\caption{Empirical initialization cost of GeoRA. We compare standard SVD and randomized SVD across model scales, and report the training cost of Qwen3-8B for reference.}
\label{tab:init_cost}
\resizebox{\columnwidth}{!}{
\begin{tabular}{l l c c}
\toprule
\textbf{Model} & \textbf{Method} & \makecell{\textbf{Wall-clock}\\\textbf{Time (min)}} & \makecell{\textbf{Peak}\\\textbf{VRAM (GB)}} \\
\midrule
8B  & Training      & 838   & $\sim$384 \\
8B  & Standard SVD  & 18.87 & 16.48 \\
8B  & Rand. SVD     & \textbf{0.21} & \textbf{15.78} \\
14B & Standard SVD  & 25.78 & 29.63 \\
14B & Rand. SVD     & \textbf{0.26} & \textbf{28.43} \\
32B & Standard SVD  & 35.23 & 64.05 \\
32B & Rand. SVD     & \textbf{0.47} & \textbf{62.38} \\
72B & Standard SVD  & 97.82 & 146.87 \\
72B & Rand. SVD     & \textbf{0.72} & \textbf{140.41} \\
\bottomrule
\end{tabular}
}
\end{table}

\begin{table*}[t]
    \centering
    \caption{Comprehensive performance comparison on Qwen2.5-1.5B and Qwen3-4B. We report In-Distribution (ID) mathematical reasoning scores and Out-of-Distribution (OOD) generalization scores. }
    \label{tab:extended_results_styled}

    \resizebox{0.8\textwidth}{!}{
    \begin{tabular}{l cccc ccc}
        \toprule
        & \multicolumn{4}{c}{\textbf{In-Distribution (ID)}} & \multicolumn{3}{c}{\textbf{Out-of-Distribution (OOD)}} \\
        \cmidrule(lr){2-5} \cmidrule(lr){6-8}
        \textbf{Method} & AIME24 & AIME25 & MATH500 & OlymMATH & HumanEval & GPQA & MMLU \\
        \midrule
        
        % --- 模型 1 ---
       \multicolumn{8}{c}{\textit{Qwen2.5-1.5B}} \\
       \midrule
        Full FT    & 7.92 & 1.25 & 53.00 & 4.00 & 45.73 & 25.10 & 55.40 \\
        SparseFT   & 8.33 & 1.67 & 53.60 & 4.50 & 51.83 & 27.50 & 59.80 \\
        LoRA       & 7.50 & 0.83 & 52.60 & 3.75 & \textbf{59.15} & 29.10 & 62.10 \\
        PiSSA      & 8.75 & 1.67 & 53.40 & 4.50 & 57.32 & 29.50 & 62.50 \\
        MiLoRA     & 9.58 & 2.08 & \textbf{54.80} & 5.25 & 58.54 & 29.80 & \textbf{62.80} \\
        \rowcolor{blue!5}
        GeoRA & \textbf{10.83} & \textbf{2.50} & 54.60 & \textbf{5.50} & 58.54 & \textbf{30.23} & 62.40 \\
        
        \midrule
        
        % --- 模型 2 ---
       \multicolumn{8}{c}{\textit{Qwen3-4B}} \\
       \midrule
        Full FT    & 12.92 & \textbf{9.58} & \textbf{73.80} & 5.00 & 3.66 & 31.46 & 61.15 \\
        SparseFT   & 12.92 & 9.17 & 72.20 & 5.25 & 4.27 & 31.96 & 63.50 \\
        LoRA       & 10.83 & 8.33 & 71.20 & 4.75 & 3.05 & 31.04 & 62.80 \\
        PiSSA      & 12.50 & 8.75 & 70.00 & 5.25 & 4.88 & 32.47 & 64.95 \\
        MiLoRA     & 11.25 & 8.33 & 71.80 & 5.25 & 4.88 & \textbf{33.22} & 64.80 \\
        \rowcolor{blue!5}
        GeoRA & \textbf{13.33} & 9.17 & 73.40 & \textbf{5.75} & \textbf{5.38} & 33.04 & \textbf{65.23} \\
        \bottomrule
    \end{tabular}
    }
\end{table*}

\subsection{Compute and Memory Complexity}

The truncated SVD on $W_{\text{geo}}$ is a one-off preprocessing step. In practice, one can use truncated or randomized SVD with time roughly linear in $r$ (e.g., $\tilde{O}(mnr)$ for dense operators), which substantially reduces the initialization overhead compared with full SVD.

GeoRA uses dense low-rank updates. Computing $\left(\frac{\alpha}{r}BA\right)x$ can be decomposed into two matrix multiplications, costing $O(rn + mr)$ per layer, which is typically more GPU-friendly than irregular sparse updates at the same parameter budget.

\subsection{Empirical Initialization Cost}

Practical overhead of SVD initialization.
Although GeoRA introduces an additional SVD-based preprocessing step, this cost is incurred only once before training. Since GeoRA only requires the top-$r$ singular components, the initialization can be accelerated with randomized SVD, whose complexity is approximately linear in $r$ for dense operators.

Table~\ref{tab:init_cost} reports the wall-clock time and peak VRAM of standard SVD and randomized SVD across model scales. For reference, we also include the training cost of Qwen3-8B under our RLVR setting. In practice, randomized SVD reduces initialization time by roughly two orders of magnitude compared with standard SVD, making initialization a sub-minute operation even at 72B scale. Moreover, this memory cost is a one-off preprocessing overhead and can be released before training.

\section{Implementation Details}
\label{app:exp_details}
\subsection{General Training Setup}

We adopt GRPO as the default RLVR optimization algorithm for all experiments. To ensure fair comparison, we keep the training data, rollout configuration, and optimization budget identical across compared methods whenever applicable. Specifically, all PEFT methods use rank $r=16$ and scaling factor $\alpha=32$, with \texttt{target\_modules} set to \texttt{all-linear}, i.e., adapters are applied to all linear layers. For GeoRA, we use sparsity ratio $\rho=0.2$ unless otherwise specified. We train all models with AdamW in bfloat16 precision on 80GB GPUs. Unless otherwise specified, the learning rate is set to $1\times10^{-6}$, the global batch size is 128, and the rollout number is 8. When KL regularization is enabled, we use a shared reference policy and set the KL coefficient to 0.001. Unless otherwise specified, all methods are trained under the same decoding configuration and differ only in the adaptation parameterization based on a single random run.
\subsection{Task-Specific Details}

\textbf{Mathematical RLVR.}
For the main mathematical RLVR experiments, we fine-tune Qwen3-8B-Base and Llama-3.1-8B-Instruct on DeepMath-103K using GRPO. In addition, we evaluate smaller-scale settings on Qwen3-4B-Base and Qwen2.5-1.5B-Instruct using GSM8K to verify whether the gains of GeoRA persist across model scales. The maximum prompt length is set to 1024 tokens and the maximum response length is set to 4096 tokens. We train on 1 node with 8 GPUs. For DeepMath-103K, we use \texttt{math-verify} for answer verification by comparing the model output against the boxed ground-truth answer under expression- and LaTeX-based extraction rules. For GSM8K, we extract the final numerical answer from the model output and compare it against the ground-truth answer using exact match. Evaluation is conducted on AIME24, AIME25, MATH500, and OlymMATH.

\textbf{Medical RLVR.}
For medical reasoning, we train Llama-3.1-8B-Instruct on AlphaMed. We use the AlphaMed training split for RL training and the corresponding test split for validation. The maximum prompt length is set to 1024 tokens and the maximum response length is set to 4096 tokens. We train on 1 node with 8 GPUs. For reward computation, we extract the final predicted option from the model output, primarily using the \texttt{\textbackslash boxed\{X\}} format and several common variants, and compare it against the ground-truth choice. A binary reward is assigned, with 1.0 for a correct answer and 0.0 otherwise. We report results on MedQA, MedMCQA, and PubMedQA.

\textbf{Coding RLVR.}
For coding experiments, we train Qwen3-32B on Eurus-2-RL-Data. The maximum prompt length is set to 1024 tokens and the maximum response length is set to 2048 tokens. We train on 2 nodes with 8 GPUs per node. For reward computation, we extract the generated code from the model output and execute it against the provided test cases in a restricted execution environment. A sample is regarded as correct only when the generated program passes the corresponding test cases within the timeout limit. During evaluation, we follow the official benchmark protocol whenever available, including the same execution environment and decoding configuration across methods. We report results on LiveCodeBench, HumanEval, and MBPP.
\begin{figure}[H]
    \centering
    \begin{subfigure}[b]{0.5\textwidth}
        \centering
        \includegraphics[width=\linewidth]{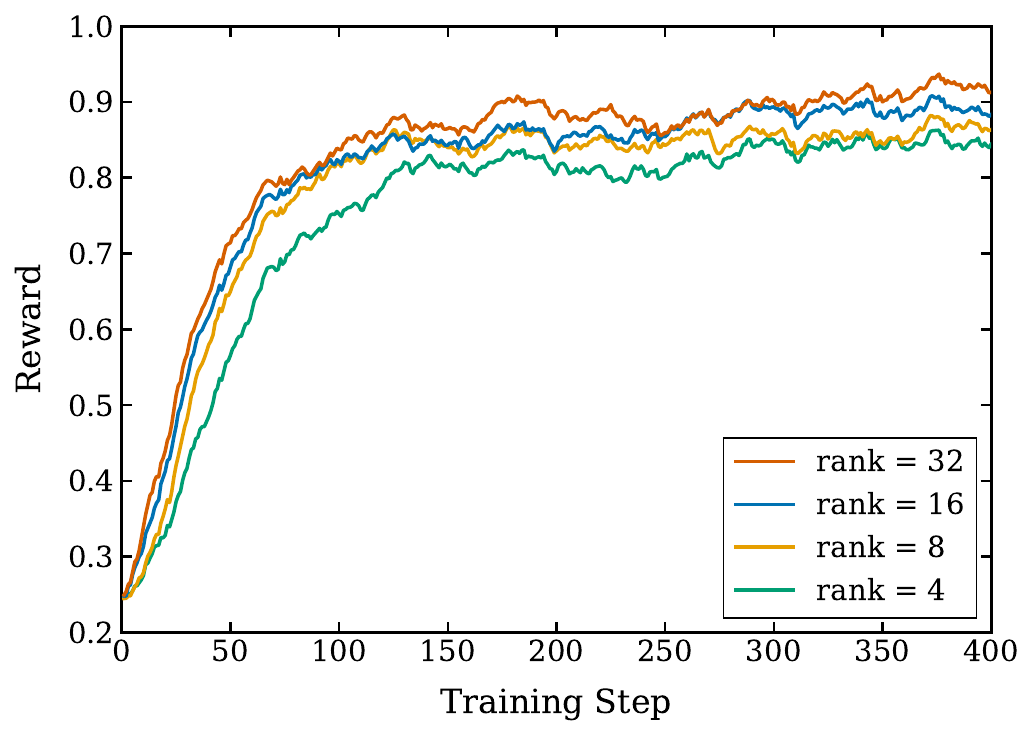} 
        \caption{Sensitivity to Rank Variations}
        \label{fig:robustness_rank}
    \end{subfigure}
    \hfill % 在两图之间通过填充空白将其撑开
    % --- 右图: Sparsity ---
    \begin{subfigure}[b]{0.5\textwidth}
        \centering
        \includegraphics[width=\linewidth]{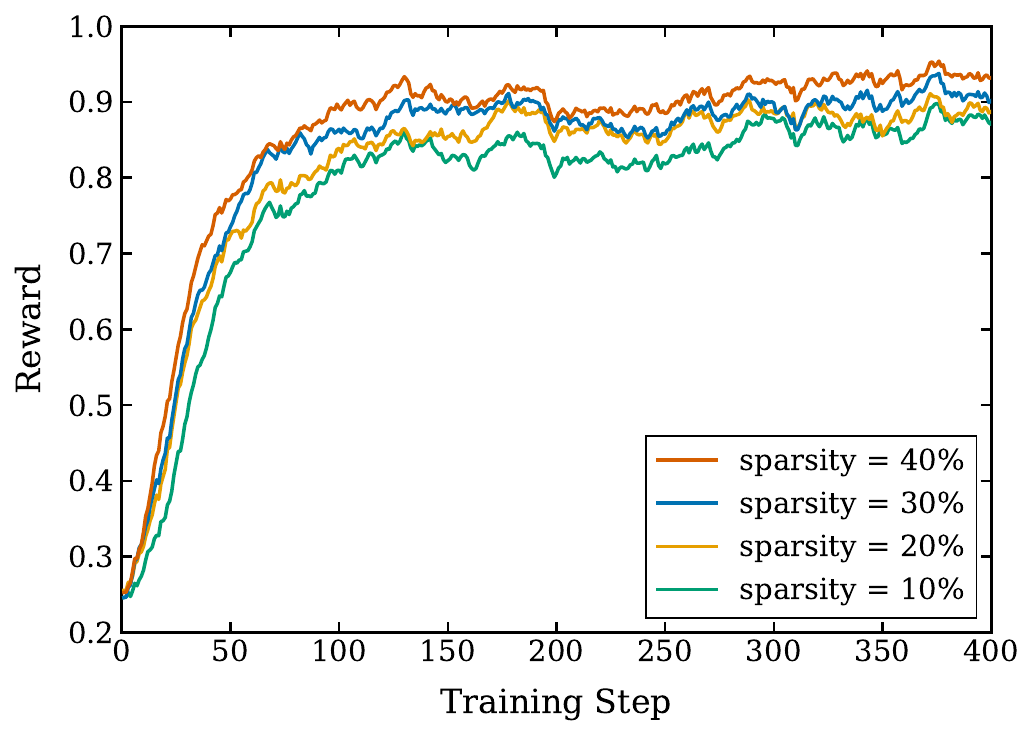}
        \caption{Sensitivity to Sparsity Variations}
        \label{fig:robustness_sparsity}
    \end{subfigure}
\caption{Parameter Robustness Analysis. }
    \label{fig:robustness}
\end{figure}

\subsection{Extended Experiments}

We further evaluate GeoRA on additional backbones with different parameter scales (Qwen2.5-1.5B and Qwen3-4B) to assess scalability and generality. Table~\ref{tab:extended_results_styled} reports both in-distribution reasoning performance and out-of-distribution generalization under the same evaluation protocol.

\section{Robustness Analysis}
\label{app:robustness}

We evaluate the sensitivity of GeoRA's performance to variations in key hyperparameters on the Qwen3-4B model (Figure~\ref{fig:robustness}). (a) Reward trajectories under varying rank $r \in \{4, 8, 16, 32\}$ (with fixed sparsity $\rho=20\%$). (b) Reward trajectories under varying sparsity levels $\rho \in \{10\%, 20\%, 30\%, 40\%\}$ (with fixed rank $r=16$). In both scenarios, GeoRA demonstrates exceptional robustness, maintaining stable convergence and consistent high reward across a wide range of parameter settings, even at extremely low ranks or high sparsity levels.

\end{document}